\documentclass[10pt,leqno]{amsart}
\usepackage{graphicx}
\usepackage{indentfirst,csquotes}
\usepackage{svg}
\usepackage{longtable}
\usepackage{booktabs}

\usepackage{amssymb,amsthm,amsmath}
\usepackage{xcolor,paralist,hyperref,fancyhdr,etoolbox} 

\hypersetup{ colorlinks=true, linkcolor=black, filecolor=black, urlcolor=black }

\newcommand{\ignore}[1]{}

\begin{document}
\title{ARDena: Scenario-Driven Control of Real-Time LLM Agents}
\author[L. Borozan]{Luka Borozan*°}
\author[D. Matijević]{Domagoj Matijević*°}
\date{\today}
\email{lborozan@mathos.hr}
\maketitle

\let\thefootnote\relax
\footnotetext{\flushleft * Faculty of Applied Mathematics and Informatics, University of Osijek Croatia, \\° Meet Intelligent Innovations LLC}

\begin{abstract}
Large language models (LLMs) have enabled increasingly capable conversational agents, but reliably controlling their behavior in real-time interactive environments remains a significant challenge. Existing approaches often rely on model fine-tuning or alignment procedures that are difficult to adapt to changing interaction requirements. This paper introduces layered scenario-driven LLM control, a framework that enables runtime behavior control through structured prompting. By combining persistent context with scenario-specific constraints, the approach allows agent behavior to be modified during interaction without changing the underlying model. The framework is implemented in ARDena, a real-time multimodal embodied agent that integrates speech interaction, visual perception, tool use, and avatar-based response generation. The proposed approach is evaluated with respect to control effectiveness, response latency, and operational stability. The results demonstrate that scenario definitions alone can produce substantially different interaction behaviors while maintaining stable real-time operation, highlighting the effectiveness of scenario-driven prompting for controlling LLM agents.
\end{abstract}

\bigskip

\section{Introduction}

Recent advances in large language models (LLMs) \cite{brown2020languagemodelsfewshotlearners, touvron2023llamaopenefficientfoundation, chowdhery2022palmscalinglanguagemodeling} have enabled the development of increasingly capable conversational agents that can reason \cite{yao2023reactsynergizingreasoningacting}, use tools \cite{schick2023toolformerlanguagemodelsteach}, and interact with users through natural language. As conversational agents and virtual assistants become increasingly integrated into everyday applications and services \cite{luger2016pa}, large language models are also being deployed within educational systems, digital characters, and embodied agents. These applications often operate in real-time interactive environments where behavior must be responsive, predictable, and aligned with specific interaction goals \cite{Wang_2024}. \\

Despite their capabilities, LLMs remain inherently stochastic systems whose behavior is not directly controlled by application logic. While modern models can follow instructions and adapt to context, reliably constraining their behavior during interaction remains a significant challenge \cite{zhang2024adaptablelogicalcontrollarge}. In practical deployments, agents are often expected to operate under task-specific objectives, dialogue rules, educational constraints, or user-dependent policies that may change dynamically over time. Traditional approaches based on model fine-tuning are costly, difficult to update, and poorly suited to scenarios where interaction requirements must be modified at runtime \cite{hu2021loralowrankadaptationlarge}. \\

Recent research has explored various approaches for controlling and extending the capabilities of large language models \cite{zhao2026survey}. Instruction tuning and reinforcement learning from human feedback improve alignment during training \cite{ouyang2022traininglanguagemodelsfollow,bai2022constitutionalaiharmlessnessai}, while prompt engineering and system prompts enable behavior shaping at inference time \cite{white2023promptpatterncatalogenhance}. More recently, agent frameworks have combined language models with external tools, memory mechanisms, and orchestration layers to support complex task execution \cite{yao2023reactsynergizingreasoningacting,schick2023toolformerlanguagemodelsteach,park2023generativeagentsinteractivesimulacra}. However, these approaches primarily focus on improving model capabilities or task performance rather than providing an explicit framework for runtime behavior control in real-time interactive environments. In particular, relatively little attention has been devoted to treating interaction logic itself as a modular and dynamically replaceable control component that can be modified during operation without changing the underlying model. \\

This raises an important question: how can the behavior of an LLM-based agent be reliably controlled in real-time interactive settings without modifying the underlying model? Addressing this problem requires a mechanism that allows behavior to be explicitly specified, inspected, and updated during operation while preserving the flexibility and general capabilities of the language model. \\

To address this challenge, we propose a control paradigm based on structured prompting, referred to as  \emph{scenario-driven LLM control}. In this approach, agent behavior is dynamically shaped during interaction through a combination of persistent contextual information and task-specific constraints, enabling runtime adaptation without modifying the underlying model. The paradigm is realized in ARDena, a real-time multimodal embodied agent that integrates speech, vision, tool use, and avatar-based interaction, demonstrating the practical applicability of the proposed approach in a real-time interactive setting. The main contributions of this work are: (1) scenario-driven LLM control, a conceptual framework for runtime control of real-time LLM agents, (2) a layered control architecture that separates persistent user context from task-specific behavioral constraints, and (3) the implementation of the framework within the ARDena system. The proposed approach is evaluated through experiments examining control effectiveness, system responsiveness, and operational stability. \\

The remainder of this paper is organized as follows. Section 2 presents the proposed framework, Section 3 describes its implementation in ARDena, Section 4 reports the evaluation results, Section 5 discusses limitations and future work, and Section 6 concludes the paper.

\section{Conceptual framework}

\begin{figure}[t]
    \centering
    \includegraphics[width=\columnwidth]{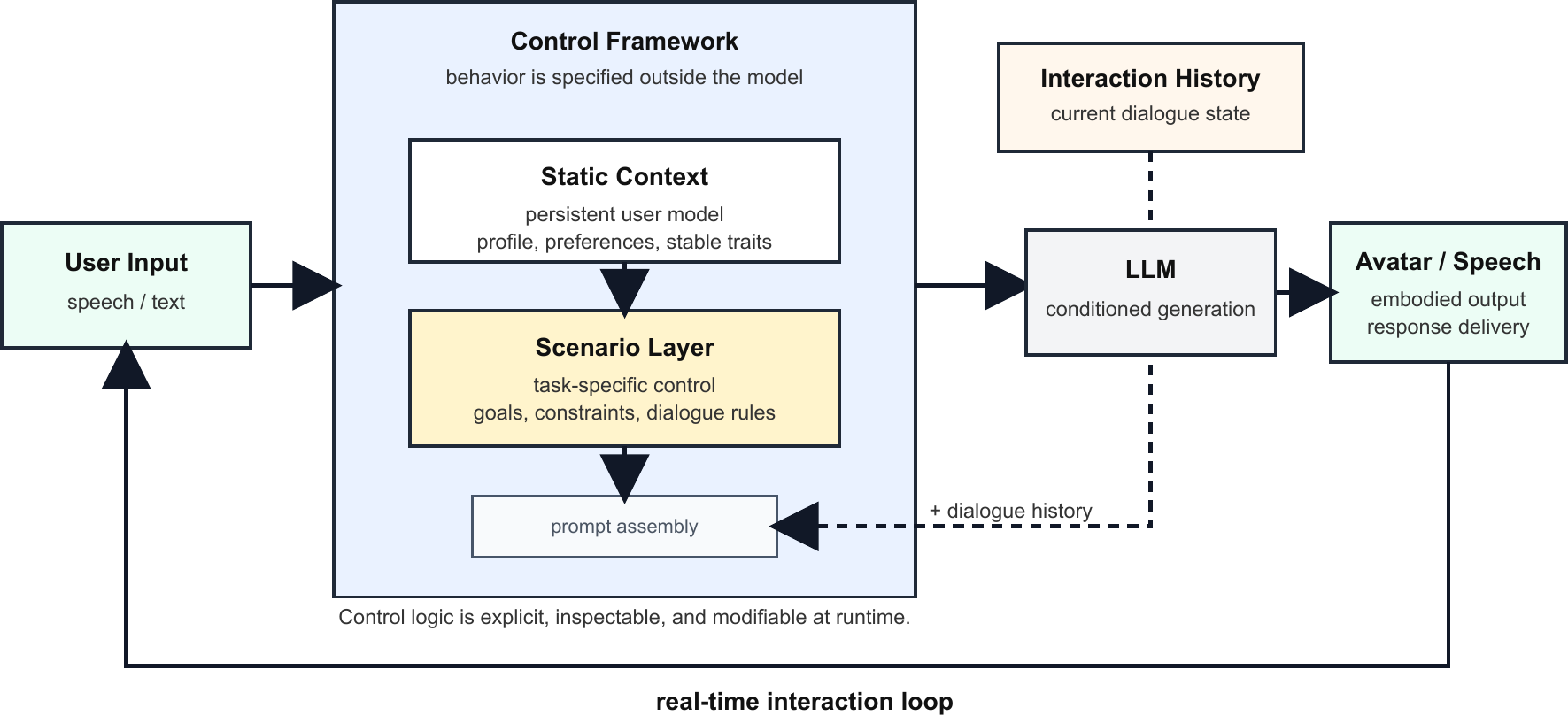}
    \caption{The proposed scenario-driven control framework. Static context, scenario-specific constraints, and interaction history are combined to guide the language model and generate real-time embodied responses.}
    \label{fig:framework}
\end{figure}

This paper introduces a framework for scenario-driven control of LLM-based agents in real-time interactive environments. The framework is motivated by the need to constrain and guide the behavior of LLMs that are inherently stochastic and not designed for structured, goal-directed interaction, through dynamic, on-the-fly constraints that are transient in nature and can be frequently updated or replaced during runtime. Rather than modifying the underlying model through fine-tuning, which is a static procedure, control is achieved at runtime through structured prompting. The proposed approach organizes control into distinct layers that jointly determine agent behavior during interaction. This enables flexible, modular, and reusable specification of interaction logic without altering model parameters. The outline of the framework can be found in Figure~\ref{fig:framework}. \\

In the next subsection, we propose the requirements the framework has to satisfy in order to qualify as a robust, real-time interactive control system for LLM-based agents. Then, we describe the control layer structure used to meet the proposed requirements. Finally, we discuss the most important aspect of the framework: the concept of scenario-driven interaction control.

\subsection{Operational Requirements}

In order to function as a real-time interactive LLM-based agent, the proposed framework must satisfy a set of operational requirements spanning model capabilities, system performance, control, personalization, and interface design explained below. \\

{\bf Model capabilities.}
The system must support structured turn-based interaction between the user and the agent. It must also handle multimodal input and output, with an emphasis on spoken interaction in real time. Furthermore, the model must maintain conversational continuity across turns and individual sessions, preserving context. Finally, it must exhibit sufficient reasoning capability to perform task-oriented inference under externally defined constraints, as well as support agentic behavior, enabling it to take initiative, manage interaction flow, and act in accordance with defined goals. \\

{\bf Performance requirements.}
Given the real-time nature of interaction, low latency and high responsiveness are essential, i.e. the system must produce outputs within time frames that do not disrupt the conversational flow. It must be robust to noise, interruptions, and imperfect input. Practical deployment requires the system to operate on common hardware. \\

{\bf Control requirements.}
A central requirement of the framework is the ability to reliably control model behavior at runtime. This includes ensuring predictability of responses, enforcing bounded behavior within predefined limits, and maintaining safety by preventing undesired or inappropriate outputs. Control mechanisms must operate without modifying model parameters, relying instead on structured input conditioning. \\

{\bf Personalization requirements.}
The system must support user-specific adaptation. This includes the ability to identify users, incorporate user-specific knowledge, and enforce individualized behavioral constraints. Additionally, the framework must allow for dynamic scenario switching, enabling runtime replacement or modification of interaction logic without interrupting the ongoing session. \\

{\bf Interface requirements.}
Finally, the system must provide clear and continuous feedback to the user. This includes visual or otherwise perceptible indicators of input capture, system processing, conversation state, and generated output. \\

The above-mentioned requirements are met through the control layers defined within our framework. These layers are described in the next subsection.

\subsection{Control Layers}


The proposed framework organizes control over the LLM into three distinct components: a static layer, a scenario layer, and an interaction loop. Together, these elements define how behavior is specified, applied, and executed during real-time interaction. The static and scenario layers operate at the level of prompt construction, while the interaction loop defines the runtime execution of the system. \\

{\bf Static Layer.} Included in every model invocation, it encompasses persistent control and information invariant across interactions. It defines the baseline behavior of the agent, including its identity, communication style, and global behavioral constraints. In addition, this layer incorporates stable user-specific context, such as user identity, biographical information, and summaries of prior interactions. Agentic capabilities are also incorporated within this layer in the form of tools, which can be defined through static prompts and invoked during interaction to extend the agent’s functionality. These elements collectively ensure continuity, personalization, and adherence to overarching safety and interaction policies. \\

{\bf Scenario Layer.} This dynamic component of the framework serves as the primary mechanism for task-level control. It is implemented as a runtime-injected instruction block that defines the immediate objective of the interaction and task-specific constraints. It may be modified or replaced during operation without interrupting the interaction. This layer is described in detail in the next subsection. \\

{\bf Interaction Loop.} It constitutes the execution layer of the framework by governing the iterative process through which user input is captured, processed, and transformed into model output. This includes handling multimodal input such as speech, managing turn-taking through voice activity detection, assembling the complete prompt from static and scenario components, invoking the language model and its agentic tools, and delivering responses through appropriate output modalities. The loop also maintains interaction state by preserving conversation history, updating summaries, and reacting to runtime events such as user changes or scenario updates. \\

The behavior of the agent emerges from the composition of static context and scenario-specific instructions, applied iteratively within the interaction loop, without requiring any modification to the underlying model parameters. In the next subsection, we will further elaborate the scenario layer and how it affects agent behavior.

\subsection{Scenarios as Dynamic Control Units}


The behavior of the agent during interaction is primarily driven by the scenario, which serves as the dynamic control unit of the proposed framework. While the static layer defines persistent constraints and baseline behavior, the scenario determines the agent’s immediate actions and responses within a given interaction context. \\

Formally, the scenario can be defined as a function that maps interaction state to constrained model behavior. It defines the type of activity being performed, such as guided dialogue, task execution, or open-ended conversation, and governs how the agent should engage with the user within that activity. In this sense, the scenario provides a goal-oriented specification of behavior that complements the global constraints imposed by the static layer. \\

At runtime, the scenario is combined with the static layer and the current interaction state to form the complete prompt presented to the language model at each iteration of the interaction loop, shaping the local context in which responses are generated. A scenario can be introduced, modified, or replaced during interaction without interrupting the system’s operation. This enables real-time adaptation of agent behavior in response to user input, performance, or external conditions. Importantly, changes to the scenario result in immediate and predictable changes in behavior, without requiring any modification to the underlying model parameters. \\

Scenarios are also inherently modular and reusable. Each scenario encapsulates a self-contained unit of interaction logic that can be applied across different users or domains with minimal modification. This allows a wide range of interaction types to be implemented through variation in scenario definitions. \\

The presented conceptual framework establishes the mechanisms that govern LLM-based agent behavior during interaction. In the following section, we demonstrate how this abstract model is concretely realized in the ARDena system, showing how its components are mapped to a practical implementation that integrates real-time processing, multimodal interaction, and scenario-based control.

\section{System Architecture and Implementation}

\begin{figure}[t]
    \centering
    \includegraphics[width=\columnwidth]{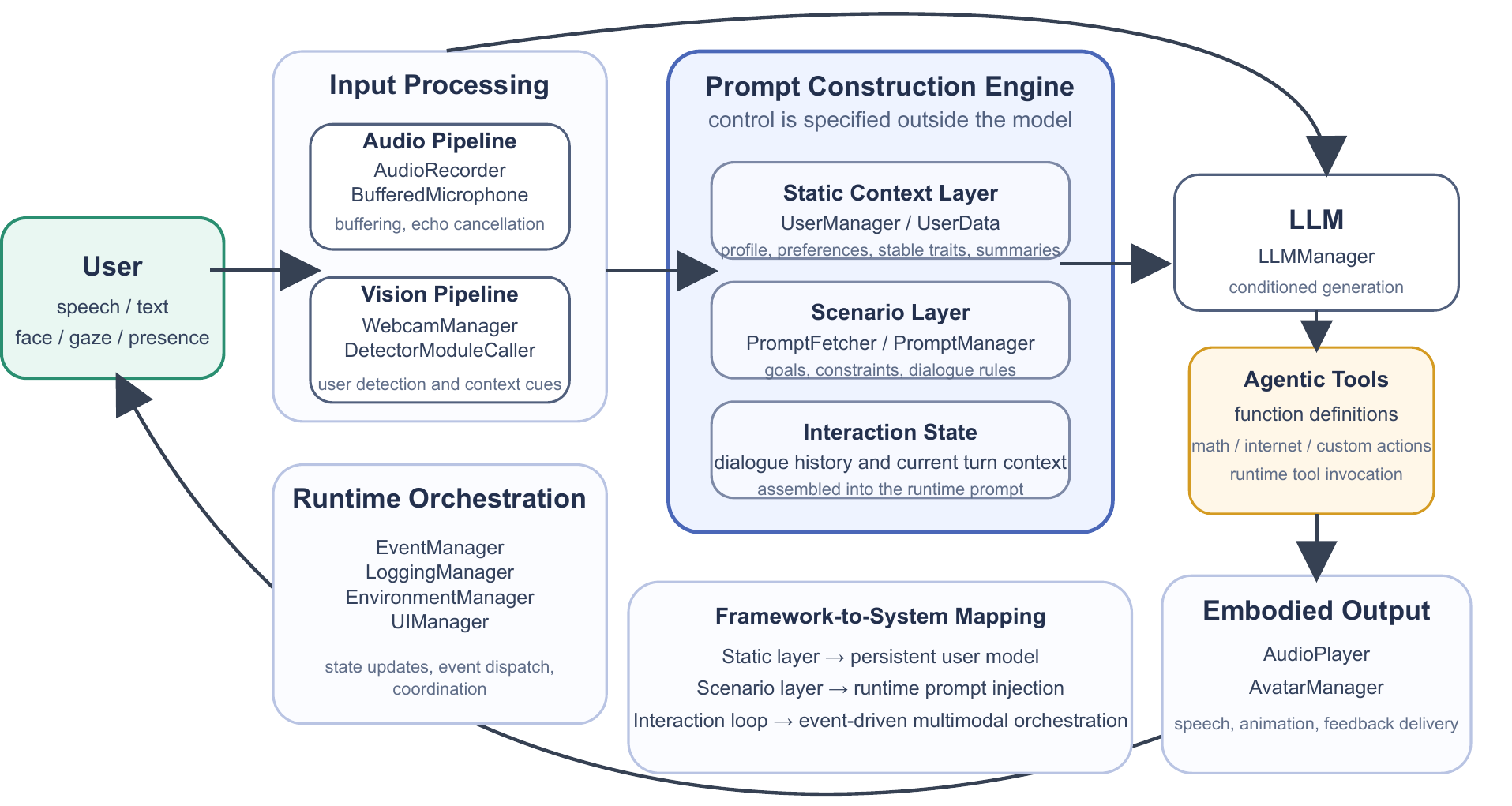}
    \caption{Architecture of the ARDena system implementing the proposed scenario-driven control framework. The diagram illustrates the flow from multimodal user input through prompt construction and LLM processing to embodied output, together with the mapping between conceptual framework components and system modules.}
    \label{fig:implementation}
\end{figure}

This section presents the architecture and implementation of ARDena, demonstrating how the scenario-driven control framework introduced in Section 2 is realized in a real-time, multimodal system. The system is implemented in C\# within the Unity engine\footnote{\url{https://unity.com/}}, enabling integration of real-time interaction, multimedia processing, and embodied output. An overview of the architecture is shown in Figure 2. At a high level, the system operates as a real-time interaction pipeline in which user input is processed and transformed into model outputs within an event-driven loop. Central to the architecture is the Prompt Construction Engine, which assembles the runtime prompt by combining persistent user context, scenario constraints, and interaction state, thereby implementing the layered control structure. \\

The remainder of this section describes the system components and their interactions, from runtime orchestration and input processing to model integration, output generation, and the overall interaction loop.

\subsection{Runtime Orchestration}
Runtime orchestration in ARDena is implemented as an event-driven infrastructure that coordinates system components and governs the flow of data during interaction. Execution is organized around a central \textit{EventManager}, which dispatches events and enables decoupled communication between modules in response to user input, scenario updates, and state changes. The \textit{EnvironmentManager} maintains shared runtime context and configuration, while the \textit{UIManager} and \textit{LoggingManager} provide real-time feedback on system state and operation. Together, these components synchronize the execution of the interaction loop, ensuring consistent behavior and low-latency responsiveness.

\subsection{Input Processing}
Input processing in ARDena handles the capture and transformation of multimodal user input into structured signals for downstream components. The system processes speech and visual input through dedicated pipelines, where audio is captured and preprocessed via the \textit{AudioRecorder}, which incorporates active echo cancellation using the WebRTC Audio Processing module\footnote{\url{https://webrtc.org/}} compiled as a native plugin. Visual input is handled by the \textit{WebcamManager}, where face detection is performed using BlazeFace \cite{bazarevsky2019blazefacesubmillisecondneuralface} and user recognition using a YOLO-based model\footnote{\url{https://docs.ultralytics.com/models/yolo11/}}, both executed through an external Python/PyTorch module. The vision pipeline provides user identification and contextual cues, enabling dynamic adaptation of system behavior. Together, these pipelines convert raw sensor data into structured representations used within the interaction loop for prompt construction and state updates.

\subsection{Prompt Construction Engine - Control Realization}
The prompt construction engine constitutes the core control mechanism of ARDena, responsible for assembling the structured input that governs LLM behavior. It combines scenario-specific constraints with the current interaction state derived from dialogue history, forming the complete runtime prompt used for model invocation. Scenario definitions are retrieved by the \textit{PromptFetcher}, which accesses an external prompt repository where prompts are authored and stored as part of predefined scenarios. Selection is driven by the currently active user, identified through the vision pipeline. The \textit{PromptManager} is responsible for organizing, formatting, and integrating the retrieved scenario content with interaction state into a coherent prompt structure for the language model. Together, these components enable explicit, dynamic control of agent behavior in accordance with the scenario-driven framework.

\subsection{Language Model and Tool Integration}
The language model is accessed through the \textit{LLMManager}, which handles model invocation and response processing during interaction. In our implementation, the realtime interaction path is handled through OpenAI Realtime models\footnote{\url{https://platform.openai.com/docs/guides/realtime}}, used for low-latency audio-to-audio turn-based dialogue, while non-realtime tasks are delegated to a separate GPT model, specifically \texttt{gpt-5.2-2025-12-11}, used for interaction history summarization and tool-related reasoning where more structured processing is required. Agentic capabilities are enabled through function calling: the realtime model issues tool calls during dialogue, after which the corresponding request is resolved through the non-realtime model and the result is returned to the realtime session. This separation preserves responsiveness in live interaction while offloading more deliberate reasoning and summarization to a higher-capability non-realtime model.

\subsection{Output Generation and Embodiment}
Output generation in ARDena is realized primarily through the realtime model’s native audio-to-audio response stream, which produces spoken output directly during interaction. Audio responses are received incrementally by the \textit{LLMManager} and forwarded to the \textit{AudioPlayer} for real-time playback with interruption support. When a function call is issued, it is resolved by the non-realtime GPT model, and the result is injected back into the realtime session, allowing the model to continue generation and present the outcome as part of the ongoing audio stream. \\

Embodiment is handled by the \textit{AvatarManager}, which controls a Reallusion CC4\footnote{\url{https://www.reallusion.com/character-creator/}} character model with standard components such as a skeleton and viseme-based facial rig. Lip synchronization is performed using SALSA\footnote{\url{https://crazyminnowstudio.com/unity-3d/lip-sync-salsa/}}, while animation state transitions are managed through the Unity Animator. Additional animations and facial expressions are created using Mixamo\footnote{\url{https://www.mixamo.com/}} and Blender\footnote{\url{https://www.blender.org/}}. The avatar is placed within a Unity scene with configured environment and lighting, and its behavior is synchronized with system events to provide coherent visual feedback during interaction.

\subsection{Interaction Loop and Framework Realization}
ARDena operates as a continuous, event-driven interaction loop in which user input is processed and transformed into model outputs in real time. At each iteration, the system captures input, updates interaction state, constructs a prompt, invokes the language model, and delivers the response, maintaining conversational continuity throughout. The prompt construction engine rebuilds the full prompt at every step by combining persistent user context, scenario-defined constraints, and the current interaction state, enabling immediate adaptation of behavior as changes in context or scenario directly influence subsequent outputs. This execution model directly realizes the conceptual framework introduced in Section~2: the static layer corresponds to persistent user context, the scenario layer to dynamically injected task constraints, and the interaction loop to the system’s event-driven orchestration. Control is therefore achieved through structured prompt composition at runtime, without modifying the underlying model. \\

Having described the implementation of the scenario-driven control framework in ARDena, we now evaluate its effectiveness in practice.

\section{Technical evaluation} 

The proposed framework was evaluated with respect to its practical suitability for real-time interactive applications. The evaluation focused on three key aspects: system performance (measured through end-to-end response latency), system stability (assessed during prolonged continuous operation), and control effectiveness (evaluated by examining the agent's ability to consistently follow diverse scenario-specific behavioral constraints). \\

The experiments were conducted on a laptop equipped with an AMD Ryzen 7 7745HX processor, an NVIDIA GeForce RTX 4060 GPU, 16 GB of DDR5 RAM, running the Windows 10 operating system. The system was tested over a 500 Mbit/s wireless network connection. ARDena was built using Unity 6000.2.13f1 and employed the OpenAI Realtime API v2 for real-time speech-to-speech interaction throughout the evaluation.

\subsection{System Performance}

To evaluate the responsiveness of the proposed framework, we conducted a series of end-to-end latency measurements during real-time interaction. The tests were performed under normal operating conditions, with latency defined as the elapsed time between the completion of a user utterance and the beginning of the agent's generated response. A total of 305 interaction instances were recorded, providing a representative sample of the performance at run time of the system. \\

\begin{figure}[t]
    \centering
    \includegraphics[width=\columnwidth]{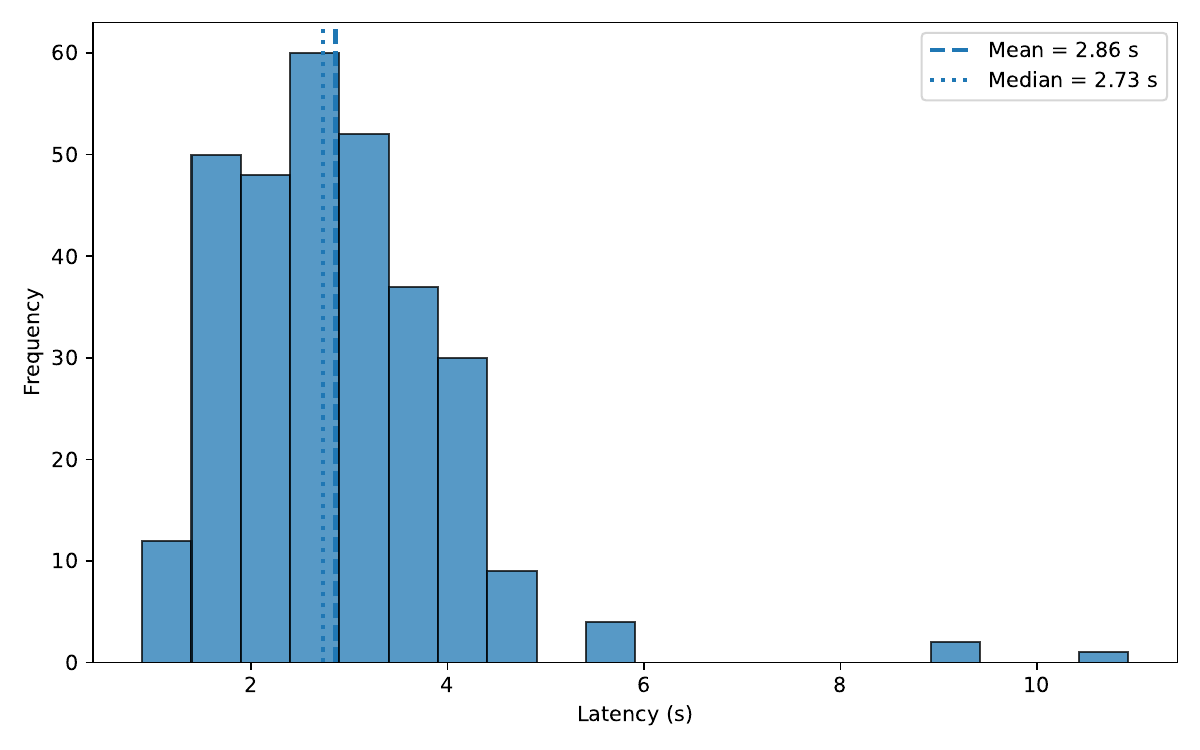}
    \caption{Distribution of end-to-end response latency measured over 305 interaction instances. The dashed and dotted vertical lines denote the mean (2.86 s) and median (2.73 s) latency, respectively.}
    \label{fig:latency}
\end{figure}

Across 305 valid latency measurements, the system achieved a mean end-to-end latency of 2.86 s and a median latency of 2.73 s, indicating that typical responses were produced within approximately three seconds. The interquartile range was 1.47 s, with the middle 50\% of measurements falling between 2.00 s and 3.48 s. Most interactions therefore remained within a conversationally acceptable range, although a small number of high-latency cases were observed. Specifically, 12.8\% of measurements exceeded 4 s, while only 2.3\% exceeded 5 s, with a maximum observed latency of 10.93 s. Note that all of the latency outliers occurred during wireless connections drops. These results suggest generally stable real-time responsiveness, with occasional outliers likely corresponding to transient processing or interaction delays. The results are depicted in Figure~\ref{fig:latency}. \\

System stability was evaluated concurrently with the latency measurements during approximately 90 minutes of continuous operation, comprising 305 conversational turns. Throughout the evaluation, no application crashes, freezes, deadlocks, or other runtime failures were observed, and no manual intervention or system restart was required. Resource utilization remained stable, with consistent memory consumption, CPU/GPU usage, while the Unity application maintained a stable frame rate. Audio input and output operated reliably throughout the tests, and the integrated echo cancellation and noise suppression mechanisms successfully prevented the system from responding to its own synthesized speech despite using the laptop's built-in speakers and microphone. Tool invocation and synchronization between the realtime interaction pipeline and auxiliary components remained consistent, and the model accurately understood user speech during the evaluation. Although three isolated latency spikes were observed, they did not interrupt the interaction or affect overall system functionality. Once the network connection was reestablished, the system automatically recovered. It is worth noting that all experiments were conducted on a mid-range consumer laptop rather than dedicated high-performance hardware, suggesting that the proposed framework can achieve stable real-time operation without requiring specialized computing resources.

\subsection{Control Effectiveness}

The primary objective of the proposed framework is to enable reliable runtime control of LLM behavior through scenario-driven prompting rather than model fine-tuning. To evaluate this capability, four substantially different scenarios were implemented and executed: a mathematics tutoring scenario, a Croatian language tutoring scenario, a guided creative storytelling scenario, and an unconstrained free conversation scenario. Each scenario imposed a distinct set of behavioral constraints, dialogue rules, and interaction objectives while relying on the same underlying language model. The evaluation focused on whether the observed agent behavior consistently followed the specified scenario instructions. In all evaluated cases, the system successfully adhered to the intended interaction logic, demonstrating that behavior could be modified by changing the scenario definition alone. The results are summarized in Table~\ref{tab:scenarios}. \\

\begin{table*}[t]
\centering
\caption{Evaluation of scenario-driven control effectiveness.}
\label{tab:scenarios}
\footnotesize
\setlength{\tabcolsep}{3pt}
\renewcommand{\arraystretch}{1.15}

\begin{tabular}{p{3.2cm}p{2.3cm}p{4.0cm}p{5.0cm}}
\hline
\textbf{Scenario instruction} & \textbf{User input} & \textbf{Expected behavior} & \textbf{Observed behavior} \\
\hline

Mathematics tutoring with predefined progression, internal state tracking, and tool-assisted verification. &
User solves arithmetic exercises and may provide incorrect answers or attempt to change the topic. &
Remain within the exercise flow, generate valid tasks satisfying the specified constraints, provide hints instead of solutions, track progress, and verify results using the mathematics tool. &
The agent consistently followed the prescribed workflow, correctly managed user responses, maintained internal progress, and successfully completed the scenario. \\

\hline

Croatian language tutoring focused on identifying nouns through multiple stages with strict dialogue rules and progress tracking. &
User identifies nouns and may provide incorrect answers or attempts to change the topic. &
Remain within the educational scenario, reject topic changes, avoid revealing solutions, provide hints after repeated failures, and advance only after correct answers. &
The agent maintained the required interaction logic, respected all constraints, and completed the scenario according to the specification. \\

\hline

Guided collaborative storytelling where the user contributes one sentence at a time and the agent extends the narrative. &
User continues the story with successive sentences or attempts to change the topic. &
Enforce the one-sentence constraint, integrate the user's contribution, extend the story coherently, and summarize the final story at the end. &
The agent successfully maintained narrative coherence, enforced the interaction rules, and generated a consistent collaborative story. \\

\hline

Unconstrained free conversation without additional task-specific restrictions. &
User discusses arbitrary topics. &
Engage naturally as a conversational assistant without educational or procedural constraints. &
The agent behaved as expected, demonstrating that changing only the scenario definition resulted in a substantially different interaction style while using the same underlying model. \\

\hline
\end{tabular}
\end{table*}

The evaluation demonstrates that the proposed scenario-driven control framework effectively constrains and shapes the behavior of an LLM agent at runtime. Despite using the same underlying model in all experiments, substantially different interaction patterns were obtained solely through changes to the scenario definition. The agent consistently adhered to scenario-specific goals, dialogue rules, and behavioral constraints without requiring model fine-tuning or parameter modification. These results support the central claim of this work that structured prompting can serve as a practical and reliable mechanism for controlling real-time LLM agents in interactive environments.

\section{Discussion and Limitations}

The results demonstrate that the proposed scenario-driven control framework provides a practical mechanism for shaping the behavior of real-time LLM agents through structured prompting and runtime policies rather than model modification. The evaluation showed that the same underlying language model could exhibit substantially different behaviors while maintaining stable operation across diverse scenarios. \\

Despite these advantages, several limitations remain. First, control is implemented through prompting rather than formal guarantees, meaning that adherence to scenario instructions depends on the language model. Although no deviations were observed in the evaluated scenarios, prompt-based control cannot guarantee perfect compliance in all situations. Second, the measured latency is influenced by external factors such as network conditions and remote model inference, making it dependent on the underlying infrastructure rather than solely on the proposed framework. Finally, the experimental evaluation considered a representative but limited set of scenarios. While these scenarios were intentionally diverse, broader studies involving additional tasks, users, and long-term deployments would provide a more comprehensive assessment of the framework's generality and robustness. \\

Future work will focus on extending the framework with more sophisticated scenario management, richer multimodal perception, adaptive scenario selection based on user state, and larger-scale user studies. Another promising direction is the integration of formal verification or policy-based mechanisms that could complement prompt-based control and provide stronger guarantees regarding agent behavior.

\section{Conclusion}

This paper introduced a scenario-driven control framework for real-time LLM agents and demonstrated its implementation through the ARDena system. Rather than modifying the underlying language model, the proposed approach achieves controllable behavior by combining persistent user context with dynamically injected scenario instructions during runtime. This layered architecture enables modular, reusable, and adaptable interaction logic while preserving the capabilities of the underlying model. The technical evaluation showed that the framework operates with low latency and stable performance on consumer-grade hardware and that the same language model can reliably exhibit substantially different behaviors across diverse scenarios through changes in the scenario definition alone. These results support the feasibility of scenario-driven prompting as a practical method for controlling real-time interactive LLM agents.

\section{Acknowledgements}

This research was supported by the European Union – NextGenerationEU through the Recovery and Resilience Facility under the Croatian National Recovery and Resilience Plan 2021–2026, project 'Ulaganje u inovativni inteligentni sustav za interakciju prirodnim jezikom u proširenoj stvarnosti - ARdena te povećavanje spremnosti za nove investicije' (NPOO.C3.2.R2-I1.04.006) and project ‘Advanced Algorithms and Optimization Models Supported by Mathematical Theory – OptimaAI’ (581-UNIOS-54).

\bigskip

\bibliographystyle{IEEEtran}
\bibliography{all}

\end{document}